\documentclass[conference]{IEEEtran}
\IEEEoverridecommandlockouts
\usepackage{cite}
\usepackage{import}
\usepackage{amsmath,amssymb,amsfonts}
\usepackage{algorithmic}
\usepackage{graphicx}
\usepackage{textcomp}
\usepackage{xcolor}
\usepackage{array}
\usepackage{multirow}
\usepackage{tikz}
\usetikzlibrary{shapes, arrows, positioning, calc, fit, arrows.meta, decorations.pathmorphing}
\tikzstyle{every picture}+=[remember picture]
\everymath{\displaystyle}
\tikzset{
	>=stealth',
	pil/.style={red!50,very thick,dotted}
}

\usetikzlibrary{positioning, fit, calc, backgrounds, shadows, arrows.meta}

\definecolor{execdark}{HTML}{1F4E79}   
\definecolor{execlight}{HTML}{DEEBF7}  
\definecolor{accentred}{HTML}{C00000}  
\definecolor{accentredlight}{HTML}{F8CBAD} 
\definecolor{accentgreen}{HTML}{A9D18E} 

\definecolor{execdark}{HTML}{1F4E79}   
\definecolor{execlight}{HTML}{DEEBF7}  
\definecolor{accentred}{HTML}{C00000}  
\definecolor{accentredlight}{HTML}{F8CBAD} 
\definecolor{accentgreen}{HTML}{548235} 
\definecolor{accentgreenlight}{HTML}{E2EFDA} 

\definecolor{violet}{RGB}{88, 66, 155}
\definecolor{darkBlue}{RGB}{0, 51, 102}       
\definecolor{darkGreen}{RGB}{0, 102, 51}      
\definecolor{darkRed}{RGB}{153, 0, 0}         
\definecolor{darkCyan}{RGB}{0, 102, 102}      
\definecolor{darkMagenta}{RGB}{102, 0, 102}   

\definecolor{orangeRed}{RGB}{237, 19, 90}

\def\BibTeX{{\rm B\kern-.05em{\sc i\kern-.025em b}\kern-.08em
    T\kern-.1667em\lower.7ex\hbox{E}\kern-.125emX}}
\begin{document}

\title{Receding-Horizon Nullspace Optimization for Actuation-Aware Control Allocation in Omnidirectional UAVs
}

\author{Riccardo Pretto$^{1}$, Mahmoud Hamandi$^{2}$, Abdullah Mohamed Ali$^{2}$\\ Gokhan Alcan$^{1}$, Anthony Tzes$^{2}$, Fares Abu-Dakka$^{2}$

\thanks{$^{1}$Riccardo Pretto and Gokhan Alcan are with the Tampere University, Finland. {\tt\small\{riccardo.pretto, gokhan.alcan\}@tuni.fi}}
\thanks{$^{2}$Mahmoud Hamandi, Abdullah Mohamed Ali, Anthony Tzes and Fares Abu-Dakka are with New York University Abu Dhabi, Abu Dhabi, UAE. {\tt\small\{mh7281, abdullah.ali, anthony.tzes, fa2656\}@nyu.edu}}
\thanks{This work was supported by the Research Council of Finland Project under Grant 370881 and has been partially supported by the NYUAD Center for Artificial Intelligence and Robotics, funded by Tamkeen under the NYUAD Research Institute Award CG010.}
}

\maketitle

\begin{abstract}
Fully actuated omnidirectional UAVs enable independent control of forces and torques along all six degrees of freedom, broadening the operational envelope for agile flight and aerial interaction tasks.
However, conventional control allocation methods neglect the asymmetric dynamics of the onboard actuators, which can induce oscillatory motor commands and degrade trajectory tracking during dynamic maneuvers.
This work proposes a receding-horizon, actuation-aware allocation strategy that explicitly incorporates asymmetric motor dynamics and exploits the redundancy of over-actuated platforms through nullspace optimization.
By forward-simulating the closed-loop system over a prediction horizon, the method anticipates actuator-induced oscillations and suppresses them through smooth redistribution of motor commands, while preserving the desired body wrench exactly. 
The approach is formulated as a constrained optimal control problem solved online via Constrained iterative LQR.
Simulation results on the OmniOcta platform demonstrate that the proposed method significantly reduces motor command oscillations compared to a conventional single-step quadratic programming allocator, yielding improved trajectory tracking in both position and orientation.
\end{abstract}

\section{Introduction} \label{sec:introduction}

Fully actuated omnidirectional multirotor platforms can independently control translational forces and rotational torques along all six degrees of freedom, allowing hovering at arbitrary attitudes and decoupled motion in constrained environments~\cite{rashad2020fully,kotarski2021performance}.
These capabilities are particularly valuable for aerial physical interaction tasks such as contact inspection, surface treatment, or tool use, where the vehicle must accurately regulate interaction forces and moments while maintaining stable flight~\cite{9462539}.

\begin{figure}[t!]
    \centering
    \input{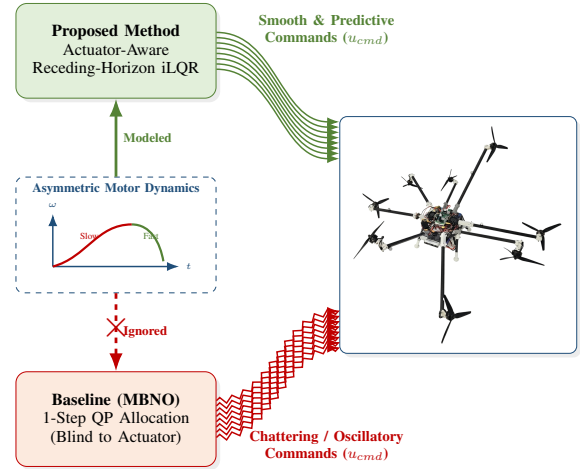}
    \caption{Conceptual overview of the proposed approach. The baseline allocator (MBNO) solves a single-step QP that is agnostic to actuator dynamics, resulting in chattering motor commands. The proposed method embeds the asymmetric motor dynamics into a receding-horizon iLQR framework, producing smooth, predictive actuator commands that anticipate and suppress oscillatory behavior.}
    \label{fig:omni_fromViser}
\end{figure}

The OmniOcta platform presented in~\cite{omniicra} achieves full actuation through eight fixed unidirectional propellers tilted at predefined angles (Fig.~\ref{fig:omni_fromViser}), avoiding servo mechanisms and relying solely on motor speed modulation to span the entire wrench space.
Although this design offers advantages in weight, mechanical simplicity, and response speed, it places the full burden of wrench generation on the motor dynamics.
Experimental characterization~\cite{omniiros} has revealed that these dynamics are notably asymmetric. The thrust rise time constant is significantly slower than the fall time constant.

Conventional control allocators, typically formulated as single-step quadratic programs~(QPs), compute actuator commands at each sampling instant without accounting for this asymmetry.
For over-actuated platforms, where infinitely many motor combinations produce the same wrench, such allocators may alternate between equivalent nullspace solutions across consecutive steps, inducing oscillatory motor commands that degrade trajectory tracking during aggressive maneuvers.

To address this limitation, we propose a receding-horizon, actuation-aware nullspace optimization strategy.
The method embeds the asymmetric motor dynamics, the rigid-body dynamics, and the feedback controller into a short-horizon optimal control problem, optimizing only the nullspace allocation variables while preserving the desired wrench.
By forward-simulating the closed-loop system, the optimizer anticipates actuator-induced oscillations and suppresses them through smooth redistribution of motor commands.

The contributions of this work are as follows:
\begin{itemize}
    \item We formulate the control allocation of an over-actuated omnidirectional UAV as a constrained optimal control problem that explicitly incorporates asymmetric motor dynamics, rigid-body dynamics, and the feedback controller over a receding prediction horizon.
    \item The proposed formulation preserves the desired wrench exactly while keeping the problem in low dimensional nullspace optimization.
    \item We validate the approach in simulation on the OmniOcta platform during maneuvers that induce oscillatory behavior under conventional single-step QP allocation, demonstrating significant reductions in motor command chattering and approximately 60\% improvement in position tracking error.
\end{itemize}

\section{Related works}
\paragraph{QP-based control allocation}
Quadratic programming is the most widely adopted approach for distributing a desired wrench among the actuators of over-actuated UAVs~\cite{oppenheimer2006control,10167718}.
At each sampling instant, a QP minimizes a quadratic cost on the actuator commands subject to bound constraints, guaranteeing instantaneous wrench feasibility.
However, since the optimization is solved independently at each time step, it has no notion of actuator dynamics or temporal consistency.
For omnidirectional platforms with a nontrivial allocation nullspace, this time-decoupled formulation may select different nullspace solutions at consecutive steps, producing discontinuous motor commands.
As reported in~\cite{omniicra}, during aggressive maneuvers such as full-flip rotations, where thrust authority shifts rapidly between intersecting propeller groups, these discontinuities induce noticeable vibrations and transient tracking degradation.
The problem is compounded by asymmetric motor dynamics~\cite{omniiros,hamandi2022static}. The differing rise and fall time constants amplify the mismatch between the commanded and realized thrust, turning allocation-level switching into sustained oscillations at the actuator output.

\paragraph{Rate-constrained and dynamic allocation}
To mitigate abrupt input changes, several authors have augmented the single-step QP with actuator-rate penalties or explicit rate constraints~\cite{Harkegard2004DynamicCA,su2022downwash}.
Although these extensions discourage large step-to-step variations in the commanded inputs, they remain single-step formulations. They penalize the rate of change at the current instant, but do not predict how the actuator state will evolve over future time steps.
Consequently, they cannot anticipate oscillations that arise from the interaction between the allocation and the asymmetric motor transient response, limiting their effectiveness during dynamic maneuvers that excite the slower rise dynamics.

\paragraph{Motor-level controllers}
An alternative line of work targets the actuator dynamics directly through dedicated motor speed controllers such as sliding-mode-based thrust tracking~\cite{7989610}.
Such approaches improve the fidelity of individual motor response but operate below the allocation layer. They do not influence how the desired wrench is distributed among redundant actuators.
Therefore, even with improved motor tracking, the allocation-level source of possible oscillatory commands remains unaddressed.

\paragraph{Model predictive control}
Model predictive control~(MPC) can, in principle, incorporate actuator dynamics and constraints over a finite prediction horizon, providing the temporal awareness that single-step methods lack~\cite{shayan2025nonlinear,probine2023model}.
However, conventional MPC formulations for multirotors typically optimize over the full state and input space, resulting in large-scale problems that are challenging to solve at the high control rates required for agile flight.
This computational burden has limited the adoption of MPC-based allocation for real-time onboard deployment on omnidirectional platforms.

\paragraph{Dynamic programming methods}
Trajectory optimization methods rooted in dynamic programming, such as Differential Dynamic Programming~(DDP) and its constrained variants~\cite{9674833,howell2019altro,ALCAN2025106220}, have demonstrated strong performance for nonlinear optimal control by iteratively refining control sequences over a finite horizon.
These methods offer favorable computational scaling for moderate-dimensional problems, making them attractive candidates for real-time deployment.
However, their application to the allocation layer of omnidirectional UAVs, where optimization can be restricted to the low-dimensional nullspace of the allocation matrix, has not been explored yet.

\medskip
The proposed method bridges this gap by formulating the allocation problem as a short-horizon constrained optimal control problem solved via Constraint iterative LQR~(CiLQR).
By restricting the optimization to the two-dimensional nullspace of the allocation matrix, the approach inherits the temporal awareness of horizon-based methods while maintaining a computational footprint compatible with high-rate online execution.

\section{Omnidirectional UAV} \label{sec:OmniUav}

An omnidirectional UAV is a fully actuated multi-rotor platform capable of independently generating forces and torques along all six degrees of freedom. 
In this work, we employ the OmniOcta platorm~\cite{omniicra}, an over-actuated omnidirectional UAV equipped with eight fixed unidirectional rotors tilted at predefined angles to achieve full wrench controllability (Fig.~\ref{fig:omni_fromViser}).
The following subsections detail the system dynamics and the control architecture adopted for the vehicle.

\subsection{System Modeling}\label{sec:systemModeling}

The state of OmniOcta is defined by position $\mathbf{p} \in \mathbb{R}^3$, linear velocity $\mathbf{v} \in \mathbb{R}^3$, attitude $R \in SO^3$ (rotation matrix from body frame to the inertial frame), and angular velocity $\boldsymbol{\omega} \in \mathbb{R}^3$. The vehicle mass is denoted by $m_R$ and its inertia matrix by $\mathbf{J}_b \in \mathbb{R}^{3 \times 3}_{+}$.
\paragraph{Dynamics}
\begin{align}
    m_R \dot{\mathbf{v}} &=
    \begin{bmatrix}
        0 \\ 0 \\ -m_R g
    \end{bmatrix}
    + R\, \mathbf{f}_B, \label{eq:trans_dynamics} \\[0.8em]
    \mathbf{J}_b \dot{\boldsymbol{\omega}} &= - \boldsymbol{\omega} \times (\mathbf{J}_b \boldsymbol{\omega}) + \boldsymbol{\tau}_B,
    \label{eq:rot_dynamics}
\end{align}
where $\mathbf{f}_B \in \mathbb{R}^3$ and $\boldsymbol{\tau}_B \in \mathbb{R}^3$ are the resultant body-frame force and moment, respectively.

\paragraph{Actuator model}

Each rotor \(i = 1,\dots,n\) is mounted at a fixed body-frame position \(\mathbf{r}_i \in \mathbb{R}^3\) and produces a thrust \(f_i\) along the unit vector \(\mathbf{u}_i\). Additionally, each rotor induces a drag torque \(\kappa_i f_i \mathbf{u}_i\) about the center of mass~(CoM), where \(\kappa_i\) is the  thrust-to-torque coefficient. The resultant body force $\mathbf{f}_B$ and moment $\mathbf{m}_B$ acting on the CoM are given by:
\begin{align}
    \mathbf{f}_B &= \sum_{i=1}^{n} f_i \mathbf{u}_i, \label{eq:body_force} \\[0.5em]
    \boldsymbol{\tau}_B &= \sum_{i=1}^{n} \left( \mathbf{r}_i \times (f_i \mathbf{u}_i) + \kappa_i f_i \mathbf{u}_i \right).
\label{eq:body_moment}
\end{align}

\paragraph{Allocation matrix}
Stacking force and moment into a generalized wrench yields 
\begin{equation}
    \begin{bmatrix}
        \mathbf{f}_B \\ \boldsymbol{\tau}
    \end{bmatrix}
    = \mathcal A \mathbf{f},
\label{eq:allocation}
\end{equation}
where $\mathbf{f} = [f_1, \dots, f_n]^\top$ and the allocation matrix $\mathcal A \in \mathbb{R}^{6 \times 8}$ has columns
\begin{equation}
    \mathcal A_{:,i} =
    \begin{bmatrix}
        \mathbf{u}_i \\
        \mathbf{r}_i \times \mathbf{u}_i + \kappa_i \mathbf{u}_i
    \end{bmatrix}.
\label{eq:allocation_column}
\end{equation}
The allocation matrix $\mathcal A$ is fully determined by the mechanical design of the vehicle~\cite{omniicra}: the tilt angles of the rotor uniquely define the thrust direction vectors $\mathbf{u}_i$, while the positions of the rotors $\mathbf{r}_i$ are fixed by construction.

\subsubsection{Motor dynamics model} \label{sec:mot_dyn_model}

An asymmetric first-order model is adopted to represent the motor dynamics, with distinct rise and fall time constants to capture the observed thrust response behavior. 
The model parameters were experimentally identified in~\cite{omniiros} and are used directly in the simulations presented in this work. 
Let $u_{\mathrm{cmd},i}(t)$ denote the input commanded to the $i$-th motor at time $t$, and let $u_{\mathrm{act},i}(t)$ denote the actual output of the corresponding motor. The motor tracking error is defined as 
\begin{align}
e_i(t) = u_{\mathrm{cmd},i}(t) - u_{\mathrm{act},i}(t-1)
\end{align}

The motor dynamics are then described by
\begin{align*}
\dot u_{\mathrm{act},i}(t) &=
\begin{cases}
\dfrac{u_{\mathrm{cmd},i}(t)-u_{\mathrm{act},i}(t-1)}{\tau_{\uparrow}}, & e_i(t)\ge 0 \quad \text{(\textbf{rise})}\\[10pt]
\dfrac{u_{\mathrm{cmd},i}(t)-u_{\mathrm{act},i}(t-1)}{\tau_{\downarrow}}, & e_i(t)< 0 \quad \text{(\textbf{fall})}
\end{cases}
\end{align*}
where $\tau_{\uparrow}$ and $\tau_{\downarrow}$ denote the rise and fall time constants, respectively. In particular, $\tau_{\downarrow} \ll  \tau_{\uparrow}$ due to the active braking capability of the electronic speed controllers on board the OmniOcta.

\begin{figure}[t!]
\centering
\input{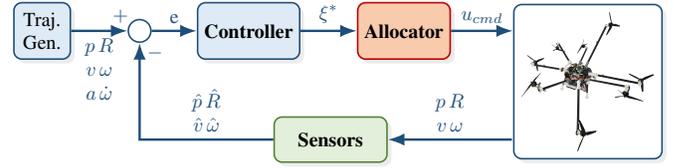}
\caption{Closed-loop control architecture of the OmniOcta. The trajectory generator provides reference states and their derivatives to the controller, which computes a desired six-dimensional wrench $\boldsymbol{\xi}^{*}$. The allocator maps this wrench to individual motor commands $\mathbf{u}_{\mathrm{cmd}} \in \mathbb{R}^8$. Onboard sensors close the loop by providing state estimates back to the controller.}
\label{fig:genericSystemScheme}
\end{figure}

\subsection{Controller Structure} \label{sec:controller_structure}

Fig. \ref{fig:genericSystemScheme} illustrates the current closed-loop control architecture of the OmniOcta, highlighting the interaction between trajectory generation, feedback control, and system dynamics.

The control pipeline starts with the reference trajectory generation module, which provides time-parameterized reference signals for position and orientation $(\mathbf{p}^r, \mathbf{R}^r) \in SO3$, as well as their respective first and second derivatives $\mathbf{v}^r, \boldsymbol{\omega}^r, \mathbf{a}^r, \dot{\boldsymbol{\omega}}^r$. 
These references are generated using seventh-order polynomials given initial and final positions and velocities over a specified duration, ensuring smooth motions with continuous velocity, acceleration, and jerk profiles.

State feedback is obtained from onboard sensors, providing estimates of position, orientation, linear velocity and angular velocity: $\hat{\mathbf{p}}, \hat{\mathbf{R}}, \hat{\mathbf{v}}, \hat{\boldsymbol{\omega}}$.
The estimated states are compared with the reference trajectory to form a tracking error signal.

The controller combines feedforward and feedback actions to compute a desired body wrench $\xi^*$, defined as 
\begin{equation}
\label{eq:desired_wrench}
\boldsymbol{\xi}^{*} = \begin{bmatrix} \mathbf{f}_w^{*} \\ \boldsymbol{\tau}_b^{*} \end{bmatrix} \in \mathbb{R}^6,
\end{equation}
where $\mathbf{f}_w^{*} \in \mathbb{R}^3$ denotes the desired force expressed in the world frame and $\boldsymbol{\tau}_b^{*} \in \mathbb{R}^3$ is the desired torque in body coordinates. 

\paragraph{Position control} \label{sec:pos_controller}
Let $\mathbf{p}^r, \mathbf{v}^r, \mathbf{a}^r \in \mathbb{R}^3$ denote the reference position, linear velocity, and acceleration from the trajectory generator. The position and velocity tracking errors are defined as
\begin{equation}
\mathbf{e}_p = \mathbf{p}^r - \hat{\mathbf{p}}, \qquad \mathbf{e}_v = \mathbf{v}^r - \hat{\mathbf{v}} .
\end{equation}
To avoid excessive control action and ensure physically realizable commands, both errors are component-wise saturated:
\begin{equation}
e_p \leftarrow \mathrm{sat}(e_p, e_{p,\max}), \qquad
e_v \leftarrow \mathrm{sat}(e_v, e_{v,\max}),
\end{equation}
where \(\mathrm{sat}(\cdot)\) denotes element-wise magnitude saturation.

An integral error term \(\mathbf{e}_{p,I}\) is maintained and updated as
\begin{equation}
\mathbf{e}_{p,I}(t+\Delta t) = \mathbf{e}_{p,I}(t) + \mathbf{e}_p \, \Delta t .
\end{equation}

The desired force in the world frame is then computed as
\begin{equation}
\label{eq:desired_force}
\mathbf{f}_w^* =
m_R (\mathbf{a}^r +g \, \mathbf{e}_3)
+ \mathbf{K}_p^p \mathbf{e}_p
+ \mathbf{K}_d^p \mathbf{e}_v
+ \mathbf{K}_i^p \mathbf{e}_{p,I},
\end{equation}
where \(m_R\) is the vehicle mass, \(g\) is the gravitational acceleration, \(e_3 = [0\;0\;1]^\top\), and \(K_p^p, K_d^p, K_i^p \in \mathbb{R}^{3\times3}\) are diagonal gain matrices. The first term acts as the feedforward~(FF) compensator, while the remaining terms constitute the feedback~(FB) correction.

\paragraph{Orientation control} \label{sec:orientation_controller}
Let \(\mathbf{R}^r \in SO(3)\) and \(\boldsymbol{\omega}^r \in \mathbb{R}^3\) denote the reference attitude and angular velocity. The orientation tracking error is defined on \(SO(3)\) as
\begin{equation}
\mathbf{e}_R = \frac{1}{2} \, \!\left( \mathbf{R}^{r,\top} \hat{\mathbf{R}} - \hat{\mathbf{R}}^\top \mathbf{R}^r \right)^\vee,
\end{equation}
where \([\cdot]^\vee\) denotes the inverse of the skew symmetric (vee) operator. The angular velocity error is
\begin{equation}
\mathbf{e}_\omega = \boldsymbol{\omega}^r - \hat{\boldsymbol{\omega}} .
\end{equation}

The desired control torque in the body frame is computed as
\begin{equation}
\label{eq:desired_moment}
\boldsymbol{\tau}_b^* =
\mathbf{K}_p^R \mathbf{e}_R
+ \mathbf{K}_d^R \mathbf{e}_\omega
+ \left[ \mathbf{J}_R \hat{\boldsymbol{\omega}} \right]_\times \hat{\boldsymbol{\omega}} ,
\end{equation}
where \(\mathbf{J}_R\) is the inertia matrix, \([\cdot]_\times\) denotes the skew-symmetric matrix operator, and \(\mathbf{K}_p^R, \mathbf{K}_d^R \in \mathbb{R}^{3\times3}_{+}\) are positive-definite gain matrices. The last term compensates for the Coriolis effect.

\paragraph{Wrench-to-motor input computation}

The desired wrench from~\eqref{eq:desired_wrench} is passed to an allocator.

The allocator maps the six-dimensional desired wrench into individual actuator commands
\begin{equation}
\mathbf{u}_{\mathrm{cmd}} \in \mathbb{R}^8,
\end{equation}

The nominal unconstrained allocation is computed as
\begin{equation} \label{eq:u_0}
\mathbf{u}_0 = \mathcal{A}_f^\dagger \, \mathbf{f}_b^\star + \mathcal{A}_m^\dagger \, \boldsymbol{\tau}_b^\star , \qquad \mathbf{u}_0 \in \mathbb{R}^8 ,
\end{equation}
where $\mathcal{A}_f \in \mathbb{R}^{3 \times 8}$ and $\mathcal{A}_m \in \mathbb{R}^{3 \times 8}$ are the force and torque sub-blocks of the allocation matrix $\mathcal{A}$ (i.e., its first three and last three rows, respectively), and $[\cdot]^\dagger$ denotes the Moore-Penrose pseudoinverse. The desired force expressed in the body frame is obtained as
\begin{equation}
\mathbf{f}_b^\star = \hat{\mathbf{R}}^\top \mathbf{f}_w^\star ,
\end{equation}
while \(\boldsymbol{\tau}_b^\star\) is the desired body torque from~\eqref{eq:desired_moment}.

\paragraph{Motor Bounds Nullspace Optimization}

Since the unconstrained allocation in~\eqref{eq:u_0} neglects the actuator limits, the resulting command \(\mathbf{u}_0\) may violate the admissible motor bounds. To enforce feasibility, a nullspace correction is introduced that exploits the redundancy of the over-actuated system without altering the desired wrench.

Let \(n_A \in \mathbb{R}^{8 \times 2}\) be a basis of the nullspace of \(\mathcal{A}\), satisfying
\begin{equation}
\mathcal{A} n_A = \mathbf{0}.
\end{equation}
Any correction along this nullspace preserves the generated wrench. The actuator command is therefore parametrized as
\begin{equation}
\mathbf{u}_{\mathrm{cmd}} = \mathbf{u}_0 + n_A \mathbf{X} ,
\label{ucmd-calc}
\end{equation}
where \(\mathbf{X} \in \mathbb{R}^2\) is the optimization variable. Following~\cite{omniiros}, $\mathbf{x}$ is obtained by solving a quadratic program that minimizes the squared norm of the control input subject to actuator bounds:
\begin{align}\label{eq:classical_QP}
\min_{\mathbf{X}\in\mathbb{R}^2}\;\frac{1}{2}\mathbf{X}^\top \mathbf{X} + (\mathbf{u}_0^\top n_A)\mathbf{X} \\
\text{s.t.} \quad  \mathbf{u}_{\min} \le \mathbf{u}_0 + n_A \mathbf{X} \le \mathbf{u}_{\max}
\end{align}
This formulation finds the closest feasible actuator command to the nominal allocation in a least-squares sense, while guaranteeing wrench consistency and respecting physical motor limits. We refer to this method as \emph{Motor Bounds Nullspace Optimization}~(\textbf{MBNO}).

In practice, both simulation and experimental results show that the \textbf{MBNO} formulation in~\eqref{eq:classical_QP} can induce oscillatory behavior in the motor commands $\mathbf{u}_{act}$, as illustrated in Fig.~\ref{fig:OC_u_act} (the corresponding simulation setup is detailed in Section~\ref{sec:simulation}).
Analysis of the simulation data (omitted here for brevity) suggests that this phenomenon originates from the asymmetric motor dynamics described in Section~\ref{sec:mot_dyn_model}: the differing rise and fall time constants create a persistent mismatch between commanded inputs $\mathbf{u}_{\mathrm{cmd}}$ and the actual outputs $\mathbf{u}_{\mathrm{act}}$.
These discrepancies amplify position and orientation tracking errors, as demonstrated in the results section. 
The effect is further exacerbated on the real platform due to additional unmodeled actuator dynamics and sensor noise.

\begin{figure*}[t!]
\centering
\input{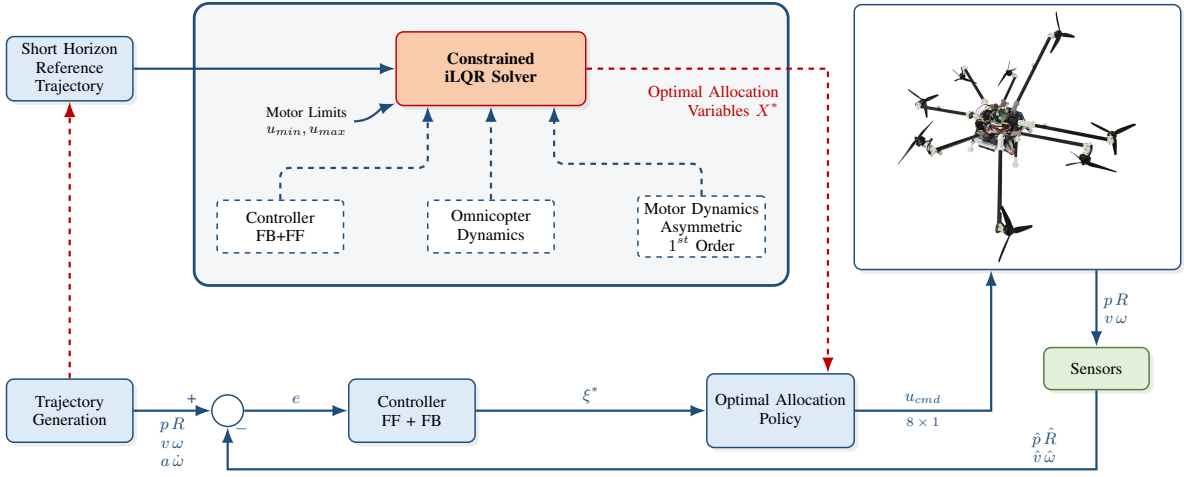}
\caption{Architecture of the proposed receding-horizon nullspace optimization. The lower loop represents the real-time control pipeline, where the optimal allocation policy maps the desired wrench to motor commands $\mathbf{u}_{\mathrm{cmd}} \in \mathbb{R}^8$. The upper block depicts the CiLQR solver, which internally forward-simulates the controller, the rigid-body dynamics, and the asymmetric first-order motor model over a short prediction horizon. By optimizing the nullspace allocation variables $\mathbf{X}^{*}$ subject to motor limits $[\mathbf{u}_{\min},\, \mathbf{u}_{\max}]$, the solver produces smooth actuator commands that are fed back to the allocation policy at each receding-horizon cycle.}
\label{fig:proposedMethodArch}
\end{figure*}

\section{Proposed Method} \label{sec:proposed_method}

To mitigate the observed chattering in $\mathbf{u}_{\mathrm{act}}$, a predictive horizon-based strategy is needed that can anticipate actuator-induced oscillations and compensate for them before they arise. 
To this end, we propose a short-horizon shooting-based MPC framework operating at the control allocation layer. 
The proposed formulation explicitly incorporates the OmniOcta rigid-body dynamics, the feedback controller structure, and the identified asymmetric actuator dynamics, while optimizing the nullspace allocation variables over a receding horizon for online execution.

By forward-simulating the closed-loop system over a prediction window, the method anticipates potential oscillations in the motor commands and proactively adjusts the nullspace variable $\mathbf{X}$ in~\eqref{ucmd-calc} to suppress them. The overall architecture of the proposed method is depicted in Fig. \ref{fig:proposedMethodArch}.

\subsection{Problem Formulation}

The state of the system at a discrete time step $k$ comprises the UAV position, velocity, orientation and angular velocity:
\begin{align}
    \mathbf{x}_k=\big[\mathbf{p}_k,\mathbf{v}_k,R_k,\boldsymbol{\omega}_k]
\end{align}
where $\mathbf{p}_k,\mathbf{v}_k \in \mathbb{R}^{3}$ are the position and the linear velocity, $R \in SO(3)$ and $\boldsymbol{\omega} \in \mathbb{R}^{3}$ is the angular velocity.

The optimization variable is the nullspace allocation parameter $\mathbf{X}_k \in \mathbb{R}^{2}$, which parametrizes the redundancy of the actuation mapping and redistributes the actuation commands among the motors.

Given the current system state $\mathbf{X}_0$ and motor output $\mathbf{u}_{\mathrm{act},-1}$ at the beginning of each planning cycle, we solve the following discrete-time constrained optimal control problem~(OCP) over a prediction horizon of $h$ steps:
\begin{subequations}\label{eq:ocp}
\begin{align}
    \min_{\mathbf{X}_{0:h-1}} \quad & \sum_{k=0}^{h-1} \ell_k\!\left(\mathbf{u}_{\mathrm{act},k},\, \mathbf{u}_{\mathrm{act},k-1}\right) \label{eq:ocp_cost}\\[0.6em]
    \textrm{s.t.} \quad
    & \mathbf{u}_{\mathrm{cmd},k}  = \mathbf{u}_{0,k} + n_A \, \mathbf{X}_k, \label{eq:ocp_alloc}\\
    & \mathbf{u}_{\mathrm{act},k}  = f_m\!\left(\mathbf{u}_{\mathrm{act},k-1},\, \mathbf{u}_{\mathrm{cmd},k}\right), \label{eq:ocp_motor}\\
    & \mathbf{x}_{k+1}  = f_r\!\left(\mathbf{x}_{k},\, \mathbf{u}_{\mathrm{act},k}\right), \label{eq:ocp_dyn}\\
    & \mathbf{0} \le \mathbf{u}_{\mathrm{cmd},k} \le \mathbf{u}_{\max}, \label{eq:ocp_bounds}\\
    & \quad k = 0, \dots, h-1. \nonumber
\end{align}
\end{subequations}
The individual components of this formulation are detailed below.

\paragraph{Nominal allocation}
At each prediction step $k$, the reference trajectory provides the feedforward and feedback terms from which the controller (Section~\ref{sec:controller_structure}) computes the desired wrench $\boldsymbol{\xi}_k^{*}$.
The unconstrained nominal allocation $\mathbf{u}_{0,k}$ is obtained from~\eqref{eq:u_0}, and the commanded input is formed via the nullspace parameterization in~\eqref{eq:ocp_alloc}.
Since the corrections are applied strictly along the nullspace of $\mathcal{A}$, the generated wrench is identical to the one requested by the controller regardless of $\mathbf{x}_k$.

\paragraph{Dynamics propagation}
The actuated motor inputs evolve according to the asymmetric first-order motor model introduced in Section~\ref{sec:mot_dyn_model}, expressed compactly in~\eqref{eq:ocp_motor}.
The rigid-body states are then propagated in~\eqref{eq:ocp_dyn} using the translational and rotational dynamics from~\eqref{eq:trans_dynamics}--\eqref{eq:rot_dynamics}, with the actual motor outputs $\mathbf{u}_{\mathrm{act},k}$ driving the system through the allocation map.
At each step, $\mathbf{u}_{\mathrm{act},k}$ is retained in memory so that the smoothness cost can be evaluated consistently.

\paragraph{Stage cost}
The stage cost function $\ell_k$ explicitly penalizes deviations in actuator commands between consecutive steps to reduce chattering:
\begin{align}
    \ell_k = \| \mathbf{u}_{\mathrm{act},k} - \mathbf{u}_{\mathrm{act},k-1} \|_{R_{\Delta u}}^2
\end{align}
where $R_{\Delta u}$ is a positive-definite weighting matrix.

\paragraph{Actuator constraints}
The element-wise inequality constraints in~\eqref{eq:ocp_bounds} enforce the physical motor limits on the commanded inputs at every prediction step, ensuring that the optimized allocation remains realizable by the hardware:
\begin{align}
    \mathbf{0} \le \mathbf{u}_{\mathrm{cmd},k} = \mathbf{u}_{0,k}+n_A \mathbf{X}_k \le \mathbf{u}_{\max}
\end{align}

\subsection{Implementation details}

\paragraph{Solver}
We employ Constrained iterative LQR (CiLQR)~\cite{howell2019altro,ALCAN2025106220} to compute feasible, locally optimal control sequences for the proposed formulation. 
CiLQR extends the classical iterative Linear Quadratic Regulator~(iLQR) to handle state and input constraints.
In standard iLQR, the optimal control sequence is obtained by iteratively linearizing the dynamics and quadratizing the cost around a nominal trajectory under the assumption of unconstrained inputs.
CiLQR incorporates constraints via an Augmented Lagrangian formulation: constraint violations are penalized in the cost function and the associated Lagrange multipliers are updated in an outer loop. 
At each iteration, an unconstrained LQR sub-problem is solved in the inner loop, after which the multipliers and penalty parameters are updated in the outer loop to progressively enforce the constraints.

\paragraph{Horizon selection}

Since the goal is to suppress oscillations caused by the mismatch between $\mathbf{u}_{cmd}$ and $\mathbf{u}_{act}$, the prediction horizon $h$ must be chosen to adequately capture the actuator transient response.
As a guideline, a first-order system reaches approximately 95\% of its setpoint within $3\tau$; accordingly, the prediction horizon is set to exceed this duration so that the optimizer can anticipate and compensate for actuator lag.

In the receding-horizon implementation, the optimization is performed over the full prediction horizon $h$, but only a short initial segment of the resulting control sequence ($h_c$) is applied before re-planning. 
Specifically, the control horizon $h_c$ corresponds to approximately the first 7\% of the prediction horizon. 
After executing the optimized commands over $h_c$ steps, the OCP~\eqref{eq:ocp} is re-solved using the updated system state.

This strategy enhances closed-loop robustness by limiting the influence of model mismatch and prediction errors that accumulate over long horizons.
Applying the optimized sequence over the entire prediction window without re-planning can lead to divergence, as inaccuracies in the forward model compound and cause the controller to overcompensate.
Frequent re-optimization with early truncation keeps the controller responsive and prevents oscillatory or unstable behavior.

\paragraph{Warm starting}
At each re-planning step, the initial guess for the optimization variables $\mathbf{X}_{0:h-1}$ is constructed from the solution of the previous cycle.  Specifically, the last applied value $\mathbf{X}_{h_c}^{*} \in \mathbb{R}^2$ from the preceding control horizon is replicated across the full prediction window and a small random perturbation is added to promote exploration:

\begin{align}
    \mathbf{X}_{0:h-1} = \mathbf{1}_{h} \otimes \mathbf{X}_{h_c}^{*}
    \;+\; \boldsymbol{\epsilon},
    \qquad \boldsymbol{\epsilon} \sim \mathcal{N}(\mathbf{0},
    \sigma^2 \mathbf{I}),
\end{align}
where $\otimes$ denotes the Kronecker product.  This warm-start strategy provides a feasible starting point close to the expected optimum, accelerating convergence and improving the numerical stability of the online solver.

\begin{figure}[t!]
    \centering
    \includegraphics[width=0.9\linewidth]{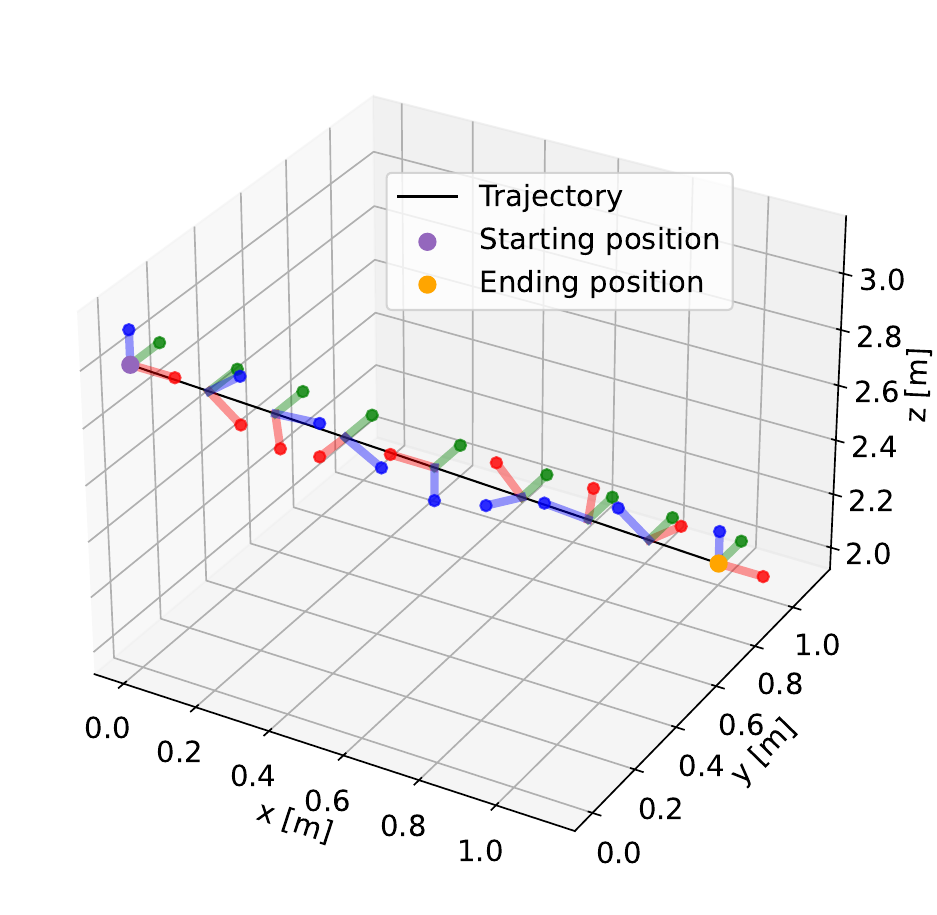}
    \caption{Reference trajectory: the OmniOcta translates from $(0,0,3)\,\mathrm{m}$ to $(1,1,2)\,\mathrm{m}$ over 60\,s while executing a full rotation about the y-axis.}
    \label{fig:ref_traj}
\end{figure}

\section{Simulation} \label{sec:simulation}

The proposed method is evaluated on a point-to-point maneuver lasting 60 seconds\footnote{The videos of the experiments can be found on the project website:\\ \texttt{https://alcanlab.github.io/rhno-uav/}}. 
The OmniOcta is commanded to translate from an inital position of $(x,y,z) = (0,0,3)$m to a final position $(1,1,2)$m, while simultaneously executing a full rotation along the y-axis.
As described in Section~\ref{sec:controller_structure}, the the reference trajectory is generated using a seventh-order polynomial to ensure continuity of velocity, acceleration, and jerk.
The resulting trajectory is shown in Fig.~\ref{fig:ref_traj}.

\begin{figure}[t!]
    \centering
    \includegraphics[width=0.9\linewidth]{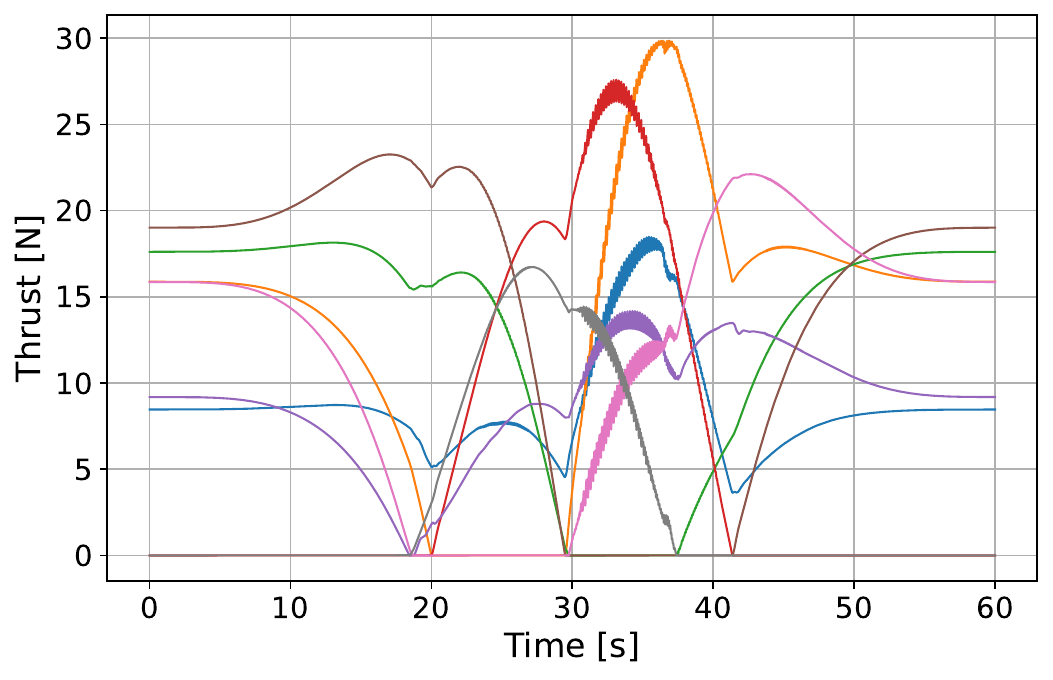}
    \caption{Actual motor commands $\mathbf{u}_{\mathrm{act}}$ under the baseline MBNO allocator. Oscillatory behavior is clearly visible between $t = 30\,\mathrm{s}$ and $t = 37\,\mathrm{s}$.}
    \label{fig:OC_u_act}
\end{figure}

\begin{figure}[t!]
    \centering
    \includegraphics[width=0.9\linewidth]{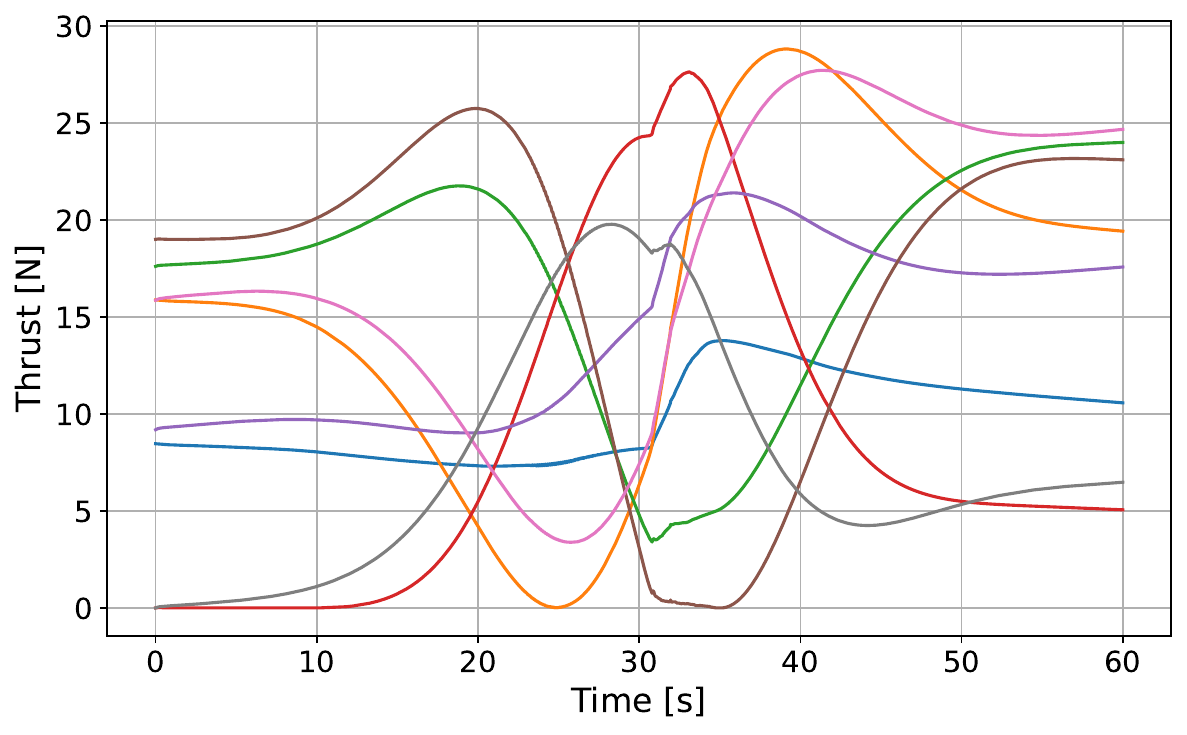}
    \caption{Actual motor commands $\mathbf{u}_{\mathrm{act}}$ over the full trajectory obtained with the proposed method. The oscillations present in the MBNO baseline (Fig.~\ref{fig:OC_u_act}) are effectively suppressed.}
    \label{fig:u_act_AAA}
\end{figure}

\begin{figure*}[h!]
    \centering
    \includegraphics[width=0.95\linewidth]{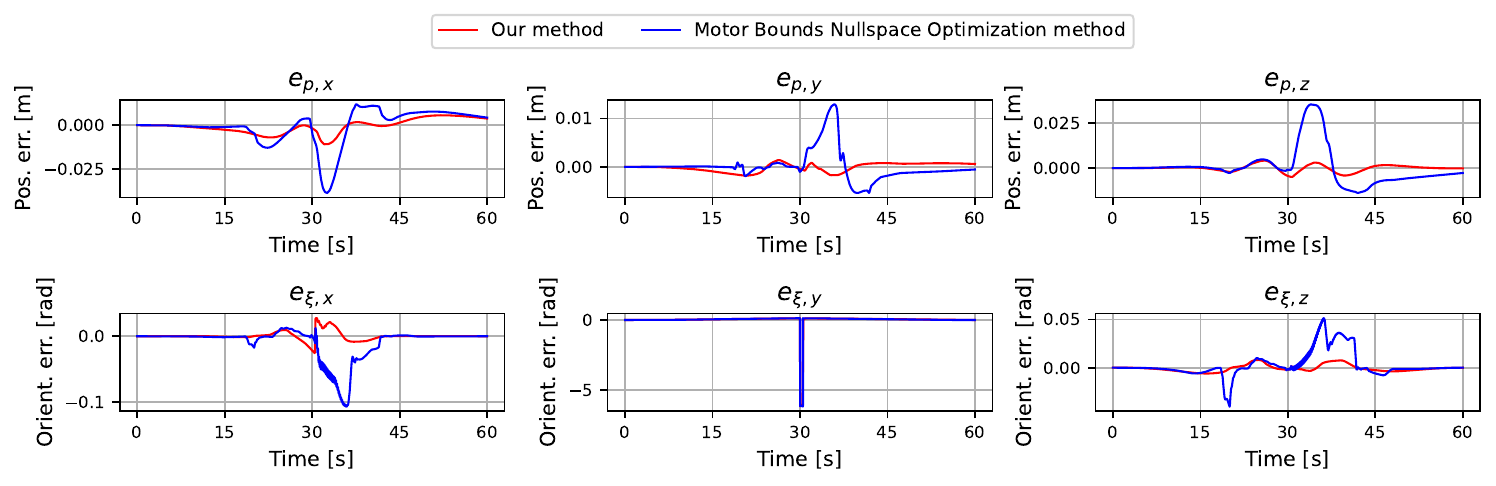}
    \caption{Position ($x$, $y$, $z$) and  orientation ($\xi_x$, $\xi_y$, $\xi_z$) tracking errors. The proposed method exhibits consistently lower error magnitudes compared to the MBNO baseline.}
    \label{fig:pos_ori_err}
\end{figure*}

\begin{figure}[t!]
    \centering
    \includegraphics[width=0.9\linewidth]{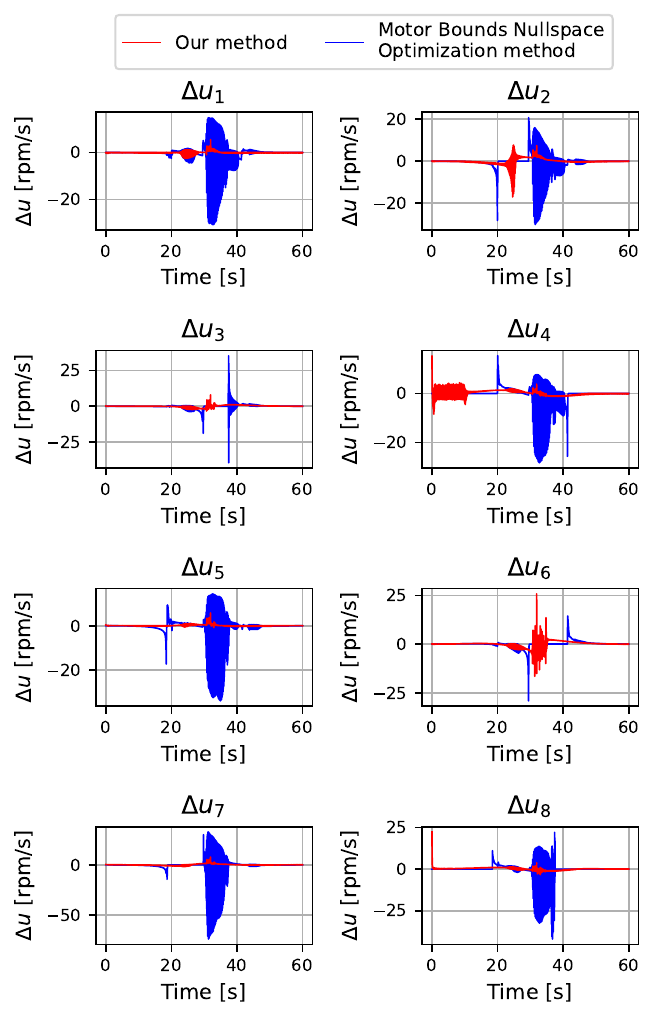}
    \caption{Comparison of actuator input variations $\Delta u$ per motor between the MBNO baseline and the proposed method. The proposed approach yields substantially smoother command profiles with reduced abrupt input changes.}
    \label{fig:delta_u_comparison}
\end{figure}

The control and sampling frequency is set to $500$Hz, corresponding to a discrete time step of $\Delta t = 0.002$s.
The prediction horizon $h$ is selected according to the guidelines in Section~\ref{sec:proposed_method} and is defined as $h = 4\bar{\tau}$, where $\bar{\tau} = \text{max}(\tau^\uparrow, \tau^\downarrow)$ accounts for the worst-case actuator time constant.
Based on the experimental identification in~\cite{omniiros}, the motor rise and fall time constants are $\tau^\uparrow = 0.15$s and $\tau_{\downarrow} = 0.021$s, respectively.
Since the dominant time constant is associated with the rise dynamics, this yields a prediction horizon of $0.6$s, corresponding to $h = 300$ discrete time steps. 

The control horizon is set to approximately 7\% of the prediction horizon, i.e., $h_c = 20$ time steps ($0.04\,\mathrm{s}$).
Shorter control horizons generally improve closed-loop performance by enabling more frequent re-optimization of the nullspace inputs. In the limiting case, applying only a single step ($h_c = 1$) before re-planning would yield the tightest feedback loop. However, this comes at the cost of solving the OCP~\eqref{eq:ocp} at every control time step. Conversely, larger control horizons were observed to degrade performance, as they reduce the controller's ability to promptly re-adapt the allocation. The chosen value of $h_c = 20$ balances tracking performance with computational feasibility.

\subsection{Results}
This section presents the performance of the proposed method compared to the baseline MBNO allocator.
The primary objective is to achieve smoother motor commands, which in turn improve trajectory tracking accuracy.

\subsubsection{Motor Command Smoothness} \label{sec:MotorInputsSmoothness}
Fig. \ref{fig:u_act_AAA} shows the actual motor outputs $\mathbf{u}_{\mathrm{act}}$ over the full trajectory obtained with the proposed method. 
Compared to the baseline MBNO results in Fig.~\ref{fig:OC_u_act}, the oscillations in the motor commands are significantly reduced.
Fig. \ref{fig:delta_u_comparison} further compares the input variation $\Delta u_{act}$ for each actuator under both methods, confirming the substantially smoother motor command profiles achieved by the proposed approach.

Around $t = 30\,\mathrm{s}$, the OmniOcta undergoes an orientation change near $\pi\,\mathrm{rad}$. Due to the discontinuity in the angular representation, the orientation wraps to $-\pi$.
This event momentarily perturbs one motor input and produces a visible slope change in Fig. \ref{fig:u_act_AAA}. 
Despite this transient, trajectory tracking remains accurate throughout.

It is also worth noting that under the proposed method, no motor drops to zero velocity during the maneuver, whereas MBNO allocator occasionally drives a pair of motors to zero.
Maintaining nonzero motor speeds is beneficial in practice, as it allows the actuators to respond more rapidly to subsequent command changes and thereby contributes to the suppression of oscillations.

\subsubsection{Trajectory Tracking}
By optimizing the nullspace variable $\mathbf{X}$ through a short-horizon, the proposed method produces more consistent actuator commands $\mathbf{u}_{act}$ while preserving the desired wrench.
The resulting reduction in oscillatory thrust redistribution mitigates excitation of the asymmetric motor dynamics and, consequently, improves trajectory tracking performance, as shown in Fig. \ref{fig:pos_ori_err}.

To quantify trajectory tracking performance, the Euclidean norms of the position and orientation errors, $\|e_p\|$ and $\|e_\xi\|$, are computed at each time step, yielding scalar time series that represent the instantaneous deviation magnitude.
The mean and root-mean-square (RMS) errors are then obtained as
\begin{equation}
\bar{e} = \frac{1}{N} \sum_{i=1}^{N} \| e_i \|, \qquad
e_{\mathrm{RMS}} = \sqrt{\frac{1}{N} \sum_{i=1}^{N} \| e_i \|^2},
\end{equation}
where $N$ is the total number of time steps.

Table \ref{tab:error_comparison} summarizes the results.
Compared to MBNO, the proposed method reduces the mean position error from $9.5$cm to $3.8$cm, and the RMS position error decreases from $14.6$cm to $4.6$cm, representing improvements of approximately 60\% and 68\%, respectively. 
Orientation errors also decrease, with the mean dropping from $0.1070$ rad to $0.1027$ rad and the RMS from $0.57$ rad to $0.5575$ rad.
These relatively large orientation error values are predominantly caused by the $2\pi$ wrapping discontinuity, which disproportionately inflates the aggregate statistics.
When considering only the portion of the trajectory prior to this switching event, the orientation errors are substantially smaller and more representative: the mean error is $0.0504\,\mathrm{rad}$ for the proposed method versus $0.0507\,\mathrm{rad}$ for MBNO, while the RMS errors are $0.0650\,\mathrm{rad}$ and $0.0657\,\mathrm{rad}$, respectively.

These results confirm that suppressing oscillations in the motor commands indirectly improves trajectory tracking, even thought the wrench requested by the controller remains unchanged.
The smoother actuator behavior enables the platform to follow the reference trajectory more faithfully.

\begin{table}[t!]
\centering
\renewcommand{\arraystretch}{1.3}
\begin{tabular}{| m{2em} | m{4em} | >{\centering\arraybackslash}m{6em} | >{\centering\arraybackslash}m{7em} |} 
\hline
 & & \textbf{Our Method} & \textbf{MBNO Method} \\ 
\hline
\multirow{2}{*}{Mean} & pos [m] & 0.0038 & 0.0095 \\ 
\cline{2-4}
                       & ori [rad] & 0.1027 & 0.1070 \\ 
\hline
\multirow{2}{*}{RMS}  & pos [m] & 0.0046 & 0.0146 \\ 
\cline{2-4}
                       & ori [rad] & 0.5575 & 0.5700 \\ 
\hline
\end{tabular}
\vspace{0.5em}
\caption{Mean and RMS tracking errors in position and orientation for the proposed method and the MBNO baseline.}
\label{tab:error_comparison}
\end{table}
\section{Conclusions and future works} \label{sec:conclusions}
This work presented a receding-horizon, actuation-aware control allocation strategy for over-actuated omnidirectional UAVs, developed and evaluated on the OmniOcta platform.
By explicitly incorporating the asymmetric motor dynamics into the optimization, the proposed method anticipates potential oscillations in the commanded motor inputs and suppresses them through smooth nullspace redistribution over a prediction horizon.

Simulation results demonstrate that the proposed approach significantly reduces actuator command oscillations compared to the baseline Motor Bounds Nullspace Optimization, yielding smoother and more consistent motor behavior.
This improvement translates into enhanced trajectory tracking, with mean position errors reduced by approximately 60\% despite the applied wrench remaining unchanged.

Future work will focus on two directions. First, we plan to address the angular representation discontinuities observed in Section~\ref{sec:MotorInputsSmoothness} by adopting a singularity-free parameterization. Second, we intend to deploy and validate the method on the physical OmniOcta platform under real-world conditions, including unmodeled dynamics and sensor noise.


\begin{thebibliography}{10}

\bibitem{rashad2020fully}
R.~Rashad, J.~Goerres, R.~Aarts, J.~B. Engelen, and S.~Stramigioli, ``Fully actuated multirotor uavs: A literature review,'' \emph{IEEE Robotics \& Automation Magazine}, vol.~27, no.~3, pp. 97--107, 2020.

\bibitem{kotarski2021performance}
D.~Kotarski, P.~Piljek, J.~Kasa{\'c}, and D.~Majeti{\'c}, ``Performance analysis of fully actuated multirotor unmanned aerial vehicle configurations with passively tilted rotors,'' \emph{Applied Sciences}, vol.~11, no.~18, p. 8786, 2021.

\bibitem{9462539}
A.~Ollero, M.~Tognon, A.~Suarez, D.~Lee, and A.~Franchi, ``Past, present, and future of aerial robotic manipulators,'' \emph{IEEE Transactions on Robotics}, vol.~38, no.~1, pp. 626--645, 2022.

\bibitem{omniicra}
M.~Hamandi, A.~M. Ali, K.~Kyriakopoulos, A.~Tzes, and F.~Khorrami, ``An omnidirectional non-tethered aerial prototype with fixed uni-directional thrusters,'' in \emph{2025 IEEE International Conference on Robotics and Automation (ICRA)}, 2025, pp. 8649--8655.

\bibitem{omniiros}
M.~Hamandi, A.~M. Ali, A.~Tzes, and F.~Khorrami, ``Experimental evaluation of safe trajectory planning for an omnidirectional uav,'' in \emph{2025 IEEE/RSJ International Conference on Intelligent Robots and Systems (IROS)}, 2025, pp. 11\,104--11\,111.

\bibitem{oppenheimer2006control}
M.~W. Oppenheimer, D.~B. Doman, and M.~A. Bolender, ``Control allocation for over-actuated systems,'' in \emph{2006 14th Mediterranean Conference on Control and Automation}.\hskip 1em plus 0.5em minus 0.4em\relax IEEE, 2006, pp. 1--6.

\bibitem{10167718}
M.~Hamandi, I.~Al-Ali, L.~Seneviratne, A.~Franchi, and Y.~Zweiri, ``Full-pose trajectory tracking of overactuated multi-rotor aerial vehicles with limited actuation abilities,'' \emph{IEEE Robotics and Automation Letters}, vol.~8, no.~8, pp. 4951--4958, 2023.

\bibitem{hamandi2022static}
M.~Hamandi, L.~Seneviratne, and Y.~Zweiri, ``Static hovering realization for multirotor aerial vehicles with tiltable propellers,'' \emph{Journal of Mechanisms and Robotics}, 2023.

\bibitem{Harkegard2004DynamicCA}
O.~H\"arkeg{\aa}rd, ``Dynamic control allocation using constrained quadratic programming,'' \emph{Journal of Guidance, Control, and Dynamics}, vol.~27, no.~6, pp. 1028--1034, 2004.

\bibitem{su2022downwash}
Y.~Su, C.~Chu, M.~Wang, J.~Li, L.~Yang, Y.~Zhu, and H.~Liu, ``Downwash-aware control allocation for over-actuated uav platforms,'' in \emph{2022 IEEE/RSJ International Conference on Intelligent Robots and Systems (IROS)}.\hskip 1em plus 0.5em minus 0.4em\relax IEEE, 2022, pp. 10\,478--10\,485.

\bibitem{7989610}
A.~Franchi and A.~Mallet, ``Adaptive closed-loop speed control of bldc motors with applications to multi-rotor aerial vehicles,'' in \emph{2017 IEEE International Conference on Robotics and Automation (ICRA)}, 2017, pp. 5203--5208.

\bibitem{shayan2025nonlinear}
Z.~Shayan, J.~Cristobal, M.~Izadi, A.~Yazdanshenas, M.~Naderi, and R.~Faieghi, ``Nonlinear model predictive control of tiltrotor quadrotors using feasible control allocation,'' \emph{Journal of Intelligent \& Robotic Systems}, vol. 111, no.~2, p.~54, 2025.

\bibitem{probine2023model}
C.~Probine, D.~Yang, K.~A. Stol, and N.~Kay, ``Model predictive control on a fully-actuated octocopter for wind disturbance rejection,'' in \emph{14th Annual International Micro Air Vehicle Conference And Competition}, 2023, pp. 11--15.

\bibitem{9674833}
G.~Alcan and V.~Kyrki, ``Differential dynamic programming with nonlinear safety constraints under system uncertainties,'' \emph{IEEE Robotics and Automation Letters}, vol.~7, no.~2, pp. 1760--1767, 2022.

\bibitem{howell2019altro}
T.~A. Howell, B.~E. Jackson, and Z.~Manchester, ``Altro: A fast solver for constrained trajectory optimization,'' in \emph{2019 IEEE/RSJ International Conference on Intelligent Robots and Systems (IROS)}.\hskip 1em plus 0.5em minus 0.4em\relax IEEE, 2019, pp. 7674--7679.

\bibitem{ALCAN2025106220}
G.~Alcan, F.~J. Abu-Dakka, and V.~Kyrki, ``Constrained trajectory optimization on matrix lie groups via lie-algebraic differential dynamic programming,'' \emph{Systems \& Control Letters}, vol. 204, p. 106220, 2025.
\end{thebibliography}
\end{document}